\documentclass{ifacconf}

\usepackage{graphicx}      % include this line if your document contains figures
\usepackage{natbib}        % required for bibliography
\usepackage{xcolor}
\usepackage{gensymb}
\usepackage{amsmath}
\usepackage{algorithm}
\usepackage{algpseudocode}
\algtext*{EndIf}% Remove "end if" text
\usepackage{url}
\usepackage{multirow}

\usepackage{fancyhdr}

\fancyhfoffset[L]{0cm} % left extra length
\fancyhfoffset[R]{0cm} % right extra length
\begin{document}
\begin{frontmatter}

\title{Optimal Economic Gas Turbine Dispatch with Deep Reinforcement Learning}

%\thanks[footnoteinfo]{Sponsor and financial support acknowledgment
%goes here. Paper titles should be written in uppercase and lowercase
%letters, not all uppercase.}

\author[First]{Manuel Sage} 
\author[Second]{Martin Staniszewski} 
\author[First]{Yaoyao Fiona Zhao}

\address[First]{Department of Mechanical Engineering, McGill University, Montreal, Canada
    (e-mail: manuel.sage@mail.mcgill.ca, yaoyao.zhao@mcgill.ca).}
\address[Second]{Siemens Energy Canada Limited, Montreal, Canada (e-mail: martin.staniszewski@siemens-energy.com)}

\begin{abstract}                % Abstract of not more than 250 words.
Dispatching strategies for gas turbines (GTs) are changing in modern electricity grids. A growing incorporation of intermittent renewable energy requires GTs to operate more but shorter cycles and more frequently on partial loads. Deep reinforcement learning (DRL) has recently emerged as a tool that can cope with this development and dispatch GTs economically. The key advantages of DRL are a model-free optimization and the ability to handle uncertainties, such as those introduced by varying loads or renewable energy production. In this study, three popular DRL algorithms are implemented for an economic GT dispatch problem on a case study in Alberta, Canada. We highlight the benefits of DRL by incorporating an existing thermodynamic software provided by Siemens Energy into the environment model and by simulating uncertainty via varying electricity prices, loads, and ambient conditions. Among the tested algorithms and baseline methods, Deep Q-Networks (DQN) obtained the highest rewards while Proximal Policy Optimization (PPO) was the most sample efficient. We further propose and implement a method to assign GT operation and maintenance cost dynamically based on operating hours and cycles. Compared to existing methods, our approach better approximates the true cost of modern GT dispatch and hence leads to more realistic policies.
\end{abstract}

\begin{keyword}
%Artificial intelligence,
reinforcement learning,
data-based control,
%power systems,
%Knowledge-based control
%Modeling and simulation of power systems,
%Control system design
%real time simulation and dispatching,
intelligent control of power systems,
optimal operation and control of power systems
\end{keyword}

\end{frontmatter}
\thispagestyle{fancy}

\lfoot{\footnotesize \textbf{\\©2023 the authors. This work has been accepted to IFAC for\\ publication under a Creative Commons Licence CC-BY-NC-ND}}
%\lfoot{\\ \footnotesize ©2023 the authors. This work has been accepted to IFAC for\\ publication under a Creative Commons Licence CC-BY-NC-ND}
%===============================================================================

\section{Introduction}
Gas turbines (GTs) play an interesting role in the discussion on modern power grids. When fired with fossil fuels such as natural gas, GTs emits greenhouse gases (GHG) and hence contribute to anthropogenic global warming. On the other hand, the significant increase in GT deployment in many areas, for example throughout North America, has allowed to phase out coal power plants which emit roughly twice as much GHG per unit of energy produced \citep{w3_coal_to_gas, w16_assessment800life}. Due to a raising incorporation of intermittent renewable energy (RE) and the lack of utility-scale energy storages for economic and technical reasons, the role of GTs in modern power grids has also changed. Fast ramping rates, partial-load operation, and higher cycle numbers are necessary to cope with the new market requirements \citep{a183_TSOUTSANIS2016} and to facilitate the integration of RE. These aspects indicate the important role that GTs play in transitioning power grids and why natural gas is sometimes termed a \textit{bridge fuel} \citep{a159_levi2013climate}. In the future, GTs fired with biogas or green hydrogen might combine the current advantages of GTs with low (or even net-zero) emissions.

The economic dispatch of GTs and hybrid energy systems (HES) involving GTs has thus been researched for many years. Mathematically, the dispatching translates into an optimization problem, and a wide range of approaches has been applied to the optimization of such systems, predominantly with the objective to minimize the cost of operation. Most approaches can be assigned to one of the following three categories \citep{a176_zhang2020}: 1) classical mathematical methods such as Lagrangian relaxation \citep{a184_ongsakul2004, a187_thorin2005}; 2) programming methods such as linear programming and dynamic programming \citep{a182_carrion, a108_ommen2014comparison}; and 3) evolutionary-inspired heuristic methods such as genetic algorithms and particle swarm optimization \citep{a185_hussain2019, a186_dixit2011}.

Recently, with reinforcement learning (RL) a fourth category has gained attention due to an increasing number of successful implementations. Inspired by the progress of deep reinforcement learning (DRL) in popular computer science control tasks such as video games or robotic simulators \citep{a79_mnih2015human, a121_schulmanPPO}, researchers have adopted DRL to the optimal dispatch problem. A key aspect behind the growing popularity of these data-driven models is that they do not require the derivation of a mathematical model of the analyzed system \citep{a178_li2020}. Model-free RL solely requires a sample model of the environment that generates the desired outputs for a set of inputs. This greatly facilitates the problem formulation and allows to incorporate existing GT models into environments for RL. Additionally, in the case of changing operating conditions, only the inputs and not the underlying formulas and constraints must be adjusted \citep{a180_zhou2020}. 

A second advantage of DRL is its ability to cope with uncertainty. It is difficult to formally capture uncertainty regarding loads and intermittent energy sources in the objective functions and constraints of traditional white-box approaches. Accounting for a large number of possible scenarios, for example by modeling longer time periods, quickly exceeds the size in which these methods can compute exact solutions \citep{a162_perera2021applications}. In the existing literature applying DRL to power dispatch, uncertainty is usually limited to RE and loads. While intermittent RE is affected stronger by uncertainty, the performance of GT also partially depends on the ambient conditions and is thus uncertain. The case study that we conduct in this work shows that throughout the course of a year the deliverable baseload power of the modeled GT can fluctuate between 22.7 MW and 30.3 MW due to varying ambient conditions. This represents a non-negligible margin of approximately 25\%.

In this study, we demonstrate the two advantages of DRL, model-free optimization and uncertainty handling, by incorporating an existing thermodynamic tool provided by Siemens Energy into our RL problem formulation. This allows a much more accurate GT power plant modeling than possible with traditional methods, taking into account the effect of ambient conditions on fuel consumption and power output. We further propose a new scheme to dynamically assign operation and maintenance (O\&M) cost, which is based on operational hours and cycle counts, to every action taken by the RL agent. Our results from a case study in Alberta, Canada show that modern DRL algorithms perform well on the GT dispatch task regarding obtained rewards and sample efficiency. Furthermore, the results indicate that the proposed O\&M cost scheme affects the policies learned and leads to more realistic GT operation.

\lfoot{}
\section{Related Work}
With soft actor-critic (SAC), a DRL algorithm was deployed in the work of \cite{a179_teng2021} to optimally dispatch an electricity-gas system. Besides a GT, the system includes a power-to-gas converter, storages for power and gas, wind turbines (WT), and photovoltaic (PV). The economic dispatch of a combined heat- and power (CHP) system via distributed proximal policy optimization (DPPO) is described in \cite{a180_zhou2020}. Here the objective is cost minimization for generating heat and power while meeting the respective demands. The DPPO algorithm is applied to both day-ahead as well as real-time dispatch. In \cite{a181_lin2020}, the DRL algorithm asynchronous advantage actor-critic (A3C) is used for the economic dispatch of a virtual power plant (VPP). The VPP consists of three subregions that are each composed of WT, PV, GT, controllable loads, and uncontrollable loads. With microgrids and multiarea integrated energy systems, similar power systems have been dispatched by hierarchical RL in the work of \cite{a177_hao2021} and \cite{a178_li2020}. Where compared to existing methods, the authors of the above studies reported better results with RL. Other studies applied RL to HES without GTs \citep{a74_phan_philippines, a91_zhang2021data}.
The various applications of RL to energy systems have been reviewed by \cite{a176_zhang2020} and \cite{a162_perera2021applications}.

In the above reviewed literature, GTs are modeled in a simplified way: via quadratic functions mapping power dispatch to operating cost \citep{a177_hao2021, a178_li2020, a181_lin2020, a180_zhou2020}, or as linear function connecting produced electricity and cost \citep{a179_teng2021, a180_zhou2020}. \cite{a179_teng2021} and \cite{a181_lin2020} also include cost associated with GHG emissions to the GT modeling approach. \cite{a181_lin2020} further expands the model with depreciation cost and O\&M cost by multiplying a cost coefficient for O\&M with the power produced by the gas turbines at every time step. Our approach differs from the related work regarding two aspects: 1) The incorporation of an existing GT modeling tool into the environment controlled by the RL agent. This enables us to replace the simple linear or quadratic functions commonly used to approximate GT operation cost. 2) The dynamic calculation of O\&M cost based on the number of GT cycles and operating hours. The approach is motivated by the observation that other implementations of O\&M cost can lead to unrealistic policies. These policies have high numbers of short cycles that would drastically increase cost in practice.

\section{Methodology}
\subsection{Problem Formulation}
For the case study, the SGT-A35, an aero-derivative GT from Siemens Energy with a nameplate capacity of 35 MW, is dispatched over a modeling period of one year with an hourly resolution. The GT is supplying power to an industrial facility with varying demands. These demands can also be met from the electricity grid at changing rates. The RL agents are thus required to learn when to operate the GT at what power level and when to purchase electricity instead. The location is assumed to be at 51.5\degree N, 111.75\degree W in Alberta. The province was chosen for three reasons: 1) it has recently experienced a transition from coal to gas and hence has significant gas infrastructure, 2) it has a large, energy-intense industry, and 3) it has great fluctuations in ambient conditions throughout the year. We use historical data from 2018 for our experiments. The climate data was obtained from the ERA5 reanalysis model \citep{era5} and consists of time-series for ambient temperature, ambient pressure, and relative humidity. For electricity, we take hourly pool prices from the \cite{aeso} and assume that the modeled power plant can purchase electricity at these rates.

\begin{figure}
	\begin{center}
		\includegraphics[width=8.4cm]{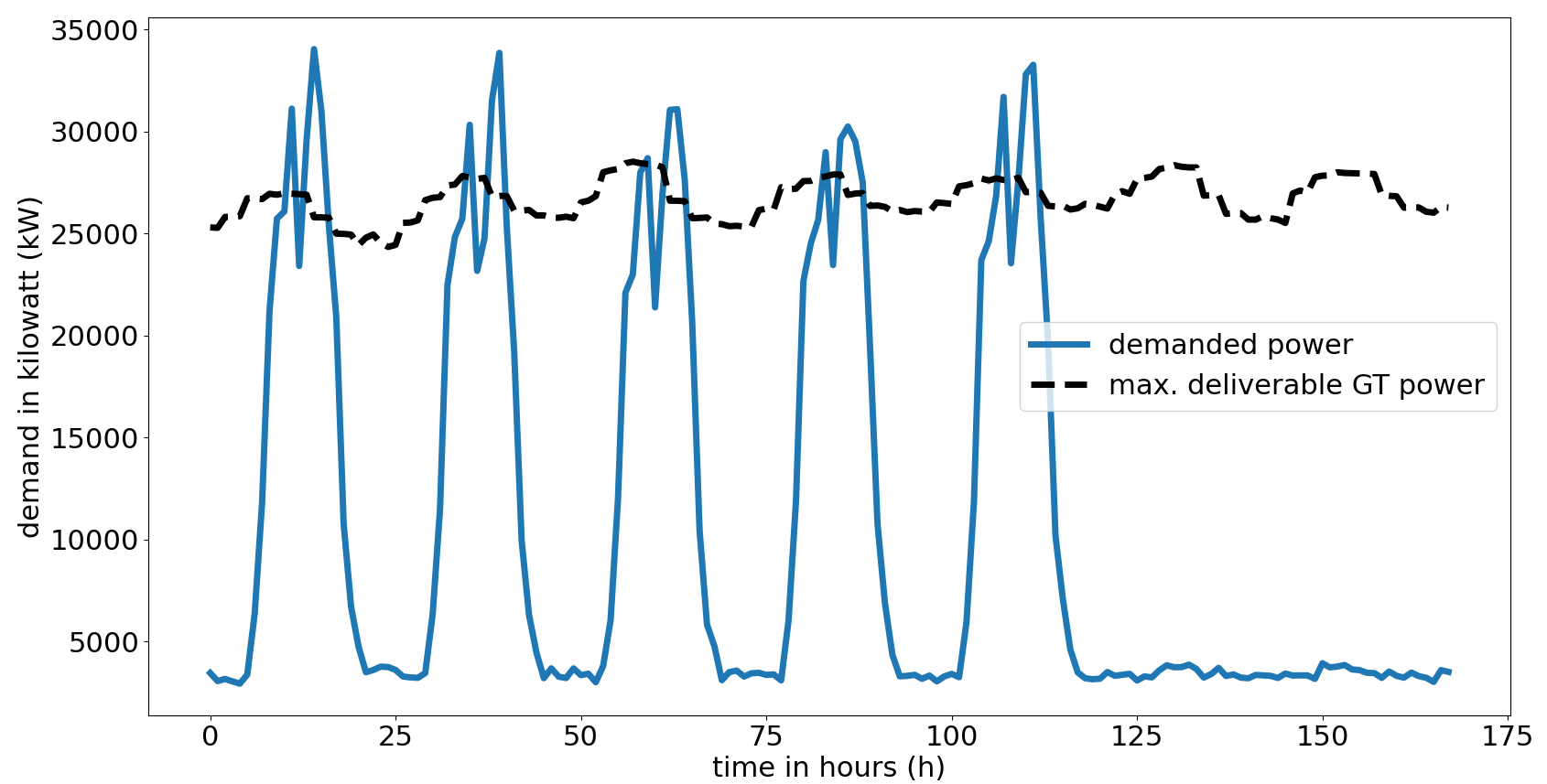}    % The printed column width is 8.4 cm.
		\caption{Example of demand profile and maximum deliverable GT power for a random week in 2018.} 
		\label{fig:demand}
	\end{center}
\end{figure}

The demand is modeled via a load profile that we synthetically generate for this case study. It mimics a shift-based demand as it occurs for manufacturing companies under single shift operation, with significant fluctuations between day and night as well as workday and weekend. The profile was generated by adding random hourly and weekly permutation, demand increases during hot and cold periods, as well as different holidays and plant-closure times to a weekly base template. Figure \ref{fig:demand} shows a random week of the obtained demand.

%The profile is in line with comparable profiles in literature [cite https://www.nature.com/articles/s41597-022-01357-8].

The dispatch of the GT is a sequential decision-making job in which the RL agent decides at what power level the GT runs for every hour. The difference between demanded and produced power is bought from the electricity grid, however, surplus electricity cannot be sold. With this choice and the varying demands, we intent to better simulate the requirements on modern GT operation regarding partial loads and cycle numbers. The goal of the agent is to minimize the total annual cost $C_{Total}$:

\begin{equation} \label{eq1}
	\min C_{Total} = \sum_{t=0}^{8760}(C_{GT}(t) + C_{Grid}(t))
\end{equation}

subject to:

\begin{equation} \label{eq2}
	C_{Grid}(t) = k_{Grid}(t) \times p_{Grid}(t)
\end{equation}
\begin{equation} \label{eq3}
	C_{GT}(t) = C_{Fuel}(t) + C_{O\&M}(t)
\end{equation}
\begin{equation} \label{eq4}
	C_{Fuel}(t) = k_{Fuel} \times q_{Fuel}(t)
\end{equation}
\begin{equation} \label{eq5}
	\begin{split}
	C_{O\&M}(t) = k_{GT,fixed} + b_{cycle}(t) \times k_{GT, cycle} \\
	+ b_{oper}(t) \times k_{GT, oper}
	\end{split}
\end{equation}
\begin{equation} \label{eq6}
	D(t) = p_{GT}(t) + p_{Grid}(t) - p_{Waste}(t)
\end{equation}
\begin{equation} \label{eq7}
	0  \leq p_{GT}(t) \leq \overline{p}_{GT}(t)
\end{equation}

\begin{figure*}
	\begin{center}
		\includegraphics[width=\textwidth]{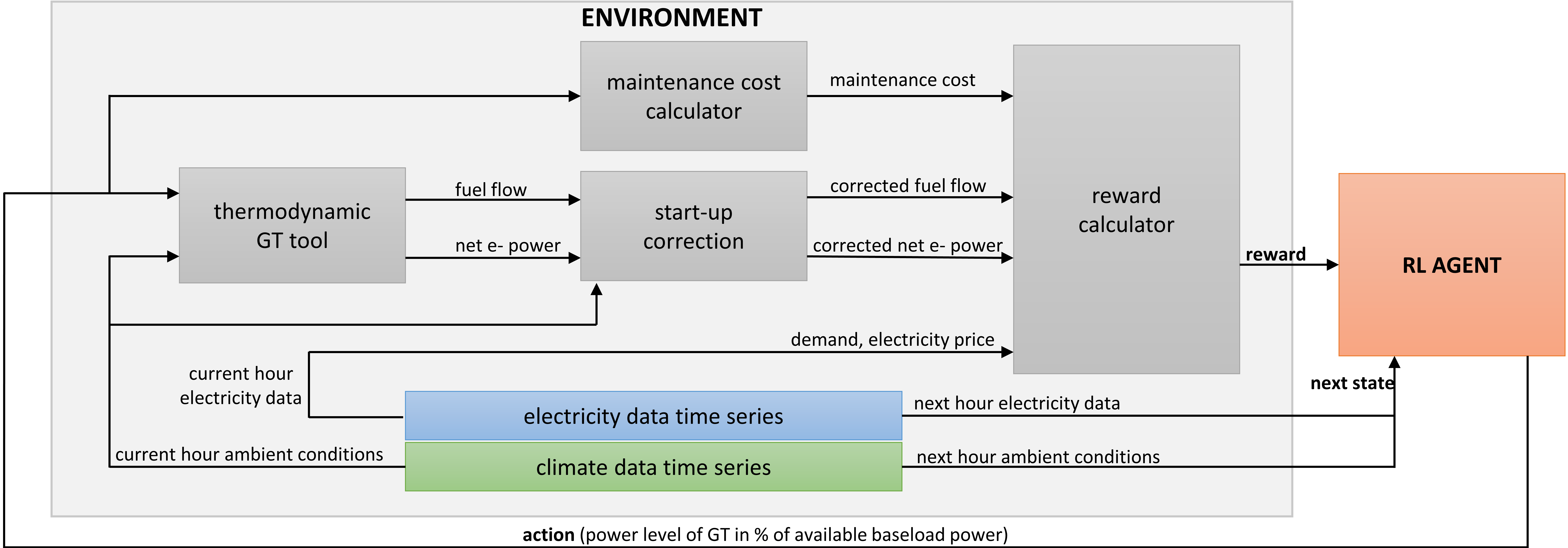} 
		\caption{Agent - environment interaction and information flow. The thermodynamic GT tool was provided by Siemens Energy. Maintenance cost, start-up correction, and reward calculator were developed for this study.} 
		\label{fig:process}
	\end{center}
\end{figure*}

Where $C_{GT} (t)$ and $C_{Grid} (t)$ represent the cost of GT usage and electricity purchase from the grid at time step $t$, respectively. $k_{Grid} (t)$ and $p_{Grid} (t)$ are the pool price and the amount of purchased electricity. The cost of GT usage is composed of the cost of fuel, $C_{Fuel}(t)$, and O\&M, $C_{O\&M}(t)$, associated with the current hour. The cost of fuel is computed by multiplying the amount of fuel burnt $q_{Fuel}(t)$ with the price of fuel $k_{Fuel}$. The fuel used for this case study is natural gas and the price is assumed to be fixed at C\$ 3.9 / GJ for the investigated period. The quantity of burnt fuel and the produced power, $p_{GT} (t)$, are determined for each hour by the provided thermodynamic tool. 
If $p_{GT} (t)$ is smaller than the industrial load, $D(t)$, the difference is bought from the grid. Excess energy is wasted ($p_{Waste} (t)$). $\overline{p}_{GT} (t)$ is the maximum deliverable power by the GT. Figure \ref{fig:process} shows how the thermodynamic tool is integrated into the environment. It is first initialized with details on the plant setup and then queried at every hour with the agent’s action and the respective ambient conditions. The tool then computes the produced net electricity and the resulting fuel flow. Researchers without access to a comparable software could replace equation \ref{eq4} by the existing approaches previously mentioned, or resort to commercial plant modeling software such as HOMER\footnote{https://www.homerenergy.com/} or Thermoflow\footnote{https://www.thermoflow.com/}.

%Where $C_{GT} (t)$ and $C_{Grid} (t)$ represent the cost of GT usage and electricity purchase from the grid at time step $t$, respectively. $k_{Grid} (t)$ and $p_{Grid} (t)$ are the pool price and the amount of purchased electricity. The cost of GT usage is composed of the cost of fuel, $C_{Fuel}(t)$, and O\&M, $C_{O\&M}(t)$, associated with the current hour. The cost of fuel is computed by multiplying the amount of fuel burnt $q_{Fuel}(t)$ with the price of fuel $k_{Fuel}$. The fuel used for this case study is natural gas and the price is assumed to be fixed at C\$ 3.9 / GJ for the investigated period. The quantity of burnt fuel and the produced power, $p_{GT} (t)$, are determined for each hour by the provided thermodynamic tool. $D(t)$ is the industrial load and $\overline{p}_{GT} (t)$ is the maximum deliverable power by the GT. Figure \ref{fig:process} shows how the thermodynamic tool is integrated into the environment. It is first initialized with details on the plant setup and then queried at every hour with the agent’s action and the respective ambient conditions. The tool then computes the produced net electricity and the resulting fuel flow. Researchers without access to a comparable software could replace equation \ref{eq4} by the existing approaches previously mentioned, or resort to commercial plant modeling software such as HOMER\footnote{https://www.homerenergy.com/} or Thermoflow\footnote{https://www.thermoflow.com/}.

The O\&M cost, $C_{O\&M}(t)$, are composed of a fixed component for each hour of the year, $k_{GT,fixed}$, a component for each GT cycle, $k_{GT,cycle}$, and a component for general operation, $k_{GT,oper}$. $b_{cycle} (t)$ is a binary variable that is 1 if the GT was started in the respective hour and zero otherwise. $b_{oper} (t)$ is a binary variable that is 1 if the GT is running for more than eight hours in a row and zero otherwise. The eight-hour threshold stems from the ratio: expected lifetime in hours over expected lifetime in cycles. Maintenance occurs after whatever is reached first, a certain number of hours or cycles. Thus, a mechanism is required that assigns cost depending on actual usage history and not only depending on the hours of operation. Algorithm \ref{algo1} shows the proposed dynamic O\&M cost calculation method in pseudocode. Note that the binary variables from Eq. \ref{eq5} have been replaced by a single non-binary variable, $GT_{state}$. This variable stores the relevant part of the operational history and is part of the state space to keep the process memoryless or Markovian.

\begin{algorithm}
	\caption{Calculation of O\&M cost, $C_{O\&M}$}
	\label{algo1}
	\begin{algorithmic}
	\State Initialize $L_{h}$, $L_{c}$, $C_{O\&M,fix}$, $C_{O\&M,var}$
	\State $GT_{state} \leftarrow 0$
	\State $GT_{hcount} \leftarrow 0$
	\For{each step $t$ of epsiode}:
		\State get action $a$ from RL agent
		\State $C_{O\&M} \leftarrow 0$
		\If{$a \neq 0$}
			\If{$GT_{state} = 0$}
				\State $C_{O\&M} \leftarrow C_{O\&M,var} / L_{c}$  \Comment{\small Cost for started cycle}
			\EndIf
			\If{$GT_{state} = 2$}
				\State $C_{O\&M} \leftarrow C_{O\&M,var} / L_{h}$  \Comment{\small Cost for long operation}
			\EndIf
		\EndIf
		\State $C_{O\&M} \leftarrow C_{O\&M} + C_{O\&M,fix} / 8760$  \Comment{\small Fixed cost}
		\If{$a = 0$}  \Comment{\small Update $GT_{state}$ and $GT_{hcount}$}
			\State $GT_{state} \leftarrow 0$
			\State $GT_{hcount} \leftarrow 0$
		\Else
			\State $GT_{hcount} \leftarrow GT_{hcount} + 1$
			\If{$GT_{hcount} \geq round(L_{h} / L_{c})$}
				\State $GT_{state} \leftarrow 2$
			\Else
				\State $GT_{state} \leftarrow 1$
			\EndIf
		\EndIf
	\EndFor
	\end{algorithmic}
\end{algorithm}

\begin{table}[hb]
	\begin{center}
		\caption{Rounded parameter values for O\&M cost calculation. GT life data and O\&M cost were estimated based on \cite{w18_turbomachinery, w19_a35_improvements} and \cite{w17_energy2020capital}, respectively.}
		\label{tb:om_params}
		\begin{tabular}{c|c|c}
			Variable name & Description & Value \\\hline
			$L_{h}$ & GT life in hours & 200,000 \\
			$L_{c}$ & GT life in cycles& 26,000 \\ 
			$C_{O\&M,fix}$ & Annual fix O\&M cost in C\$ & 780,000 \\ 
			$C_{O\&M,var}$ & Lifetime variable O\&M cost in C\$ & 33,000,000 \\ 
		\end{tabular}
	\end{center}
\end{table}

Table \ref{tb:om_params} shows approximate values for the O\&M parameters. Rounded to the nearest integer, the ratio of lifetime in hours over lifetime in cycles is eight. In other words, if the average cycle has a duration of around eight hours, cycle and hour-based cost contribute equally to the variable O\&M cost. The traditional, purely time-based cost estimation method fails to assign realistic cost for shorter cycle times. For example, with average cycles of two hours, the variable lifetime O\&M cost would still be assigned throughout 200,000 hours but actually occur within only 52,000 hours when 26,000 cycles are reached. The proposed method solves this shortcoming by assigning cycle-based O\&M cost whenever the GT is started, and hour-based O\&M cost for each hour in cycles longer than eight hours.

The last addition to the environment is a start-up correction. When the GT is running, changes between power levels only take a few minutes due to high ramp rates. This does not affect hourly production considerably and is thus neglected. Starting the GT, however, has a profound impact on the current hour which is furthermore dependent on the ambient temperature. A regular start of the modeled A35 variant takes around 20 min from turn on to baseload power, while a start below 0\degree C requires an additional 15 min of engine warm-up, called mechanical idle. For the sake of simplicity, we assume the fuel consumption during start-up is equal to the fuel consumption at mechanical idle and apply a correction to the computed hourly net electricity and fuel flow depending on the temperature. This reduces the reward in hours when the GT is started and makes short cycles less attractive for the RL agent.

\subsection{Reinforcement Learning}
RL is a subcategory of machine learning in which an agent learns by interacting with the environment. As a feedback, the agent obtains rewards for the interactions and seeks to maximize the accumulated episodic rewards, i.e. the sum of rewards over one episode. In our experiments, the reward is the negative of the cost function at every hour and one episode is one year (see Eq. \ref{eq1}). The state vector is 6-dimensional and storing the following variables: pool price, load, ambient temperature, ambient pressure, relative humidity, and GT-state. The action space is one-dimensional and bounded to range [0,1] by the environment, representing 0-100\% of the available GT power at the respective hour. The implemented algorithms are REINFORCE, deep Q-networks (DQN), and proximal policy optimization (PPO). For REINFORCE, a continuous and a discrete version are tested. Together with DQN (discrete) and PPO (continuous), this results in two algorithms for each category. The discrete algorithms choose from seven possible actions: [0, 0.5, 0.6, 0.7, 0.8, 0.9, 1] which are unevenly spaced due to significantly decreasing efficiencies under partial loads.

The DRL algorithms deployed in this study differ regarding their category and complexity. We implement all algorithms in PyTorch with artificial neural networks (ANNs) as function approximators. REINFORCE \citep{a188_williams1992simple} is a policy-gradient Monte-Carlo algorithm. Implemented with a single policy network without baseline, it is the simplest DRL algorithm tested. PPO \citep{a121_schulmanPPO} is a more sophisticated policy gradient algorithm with a modern actor-critic architecture. PPO seeks to keep policy updates within an approximated trust-region by optimizing a clipped surrogate loss. This algorithm has performed particularly well on continuous control tasks. PPO and REINFORCE train stochastic policies where actions are obtained by sampling from a learned normal distribution.
DQN \citep{a79_mnih2015human} belongs to the group of value-iteration algorithms and its update is based on tabular Q-learning. Further modifications such as experience replay and a periodically updated target network enabled this algorithm to outperform existing methods on discrete control tasks when it was introduced. Both PPO and DQN are among the most popular DRL algorithms and commonly used to benchmark state-of-the-art methods. Our implementations of PPO and DQN are based on code from \cite{software_ppo} and \cite{software_dqn}, respectively. For detailed discussions of the three algorithms, the reader is referred to the original papers.

We compare the results of the four tested variants to two baselines: the cross-entropy method (CEM) and a rule-based baseline. The CEM, introduced by \cite{a189_rubinstein1999cross}, is a heuristic, population-based optimization method  that often serves as baseline for continuous RL algorithms. Our implementation of CEM is based on the work of \cite{software_cem}. The rule-based baseline is implemented via a single if-then-else statement. If the condition is true, the GT is operated at baseload power (100\%), else it is turned off. Among the state-variables, demand and electricity price influence the reward the most. The conditions for the if-statement are thus different thresholds for these two variables and possible logical combination of them via AND and OR statements. These simple rules don’t allow for operation on partial loads, but instead operate the GT always at highest efficiency.

\section{Results and Discussion}

\begin{figure}
	\begin{center}
		\includegraphics[width=8.4cm]{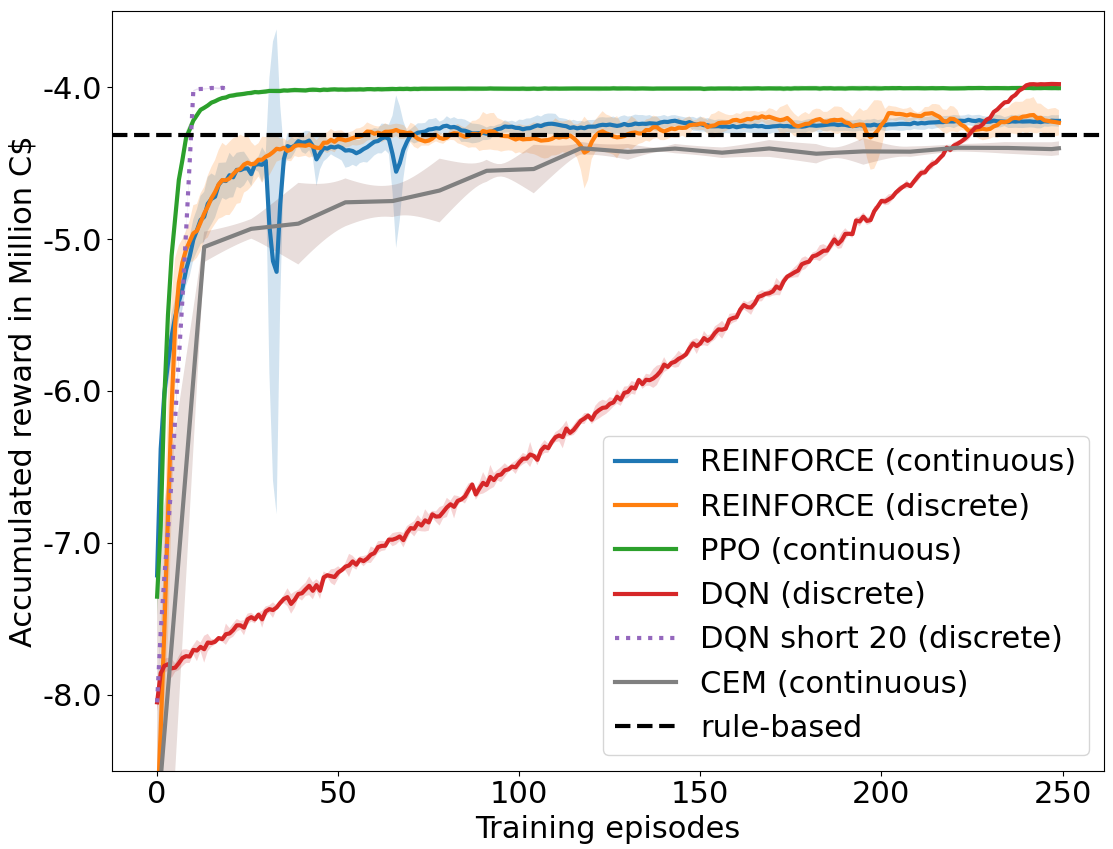}    % The printed column width is 8.4 cm.
		\caption{Training curves of implemented models.} 
		\label{fig:performance}
	\end{center}
\end{figure}

We perform hyperparameter tuning for each algorithm and report the best results obtained regarding reward maximization and sample efficiency. All results reported in this section are average values over five independent runs. Fig. \ref{fig:performance} shows the training curves of the algorithms for a training time of 250 episodes. The shaded area represents +/- one standard deviation and allows to assess the stability of the tested models. The best rule found for the rule-based baseline scored C\$ -4.32 million and operates the GT if the pool price exceeds C\$ 95/MWh. All four DRL algorithms outperform this baseline and learn better GT dispatch policies than the CEM. Both REINFORCE variants behave similarly and converge to lower episodic rewards while showing greater fluctuations compared to DQN and PPO. The training curve for DQN is highly dependent on the selected exploration-exploitation strategy. We use epsilon-greedy action selection with linear epsilon-decay from 0.8 to 0.001. To obtain comparable results, we keep epsilon fixed at 0.001 for the last 10 episodes and report averaged episodic rewards over 5 runs and 10 episodes in Table \ref{tb:performance}. With annual cost of C\$-3.98 million, DQN reached the overall best rewards.

Sample efficiency is computed in a similar fashion by averaging the rewards over the first 20 episodes. Again, the choice of epsilon significantly influences this metric for DQN. We notice that only a few episodes are required for DQN to converge. Thus, its sample efficiency is measured for a shorter training period of 20 episodes, 10 with decaying epsilon and 10 with fixed epsilon. The dotted line in Fig. \ref{fig:performance} shows this experiment. Sample efficiency scores for all algorithms are listed in Table \ref{tb:performance}. PPO is the most sample efficient algorithm, CEM again performs the worst in this metric. 

Besides analyzing the performance, we investigate the effect of the proposed dynamic O\&M cost scheme on the learned policies. Therefore, PPO and DQN are retrained over 50 episodes using three different variants of the environment: one with the proposed dynamic scheme, one with O\&M cost based on operational hours only (i.e., $b_{cycle}$ in Eq. \ref{eq5} is always 0 and $b_{oper}$ is 1 if the GT operates and 0 otherwise), and one with no variable O\&M cost (i.e., $b_{cycle}$ and $b_{oper}$ are always 0).

We compare the environments regarding average rewards, annual operational hours, and annual cycles of the GT as displayed in Table \ref{tb:results_overhaul}. Interestingly, despite obtaining nearly the same rewards, the two algorithms differ in the learned policies. DQN dispatches the GT more often, causing both higher annual cycle and hour counts. A clear tendency across both algorithms is visible for the different variants of O\&M cost calculation. Averaged over both algorithms, purely hourly-based O\&M costs led to operating time increases of 32\% in hours and of even 95\% in cycles. If no variable O\&M costs are assigned, the increase is even higher: 95\% for operating hours and 198\% for cycles compared to the proposed cost scheme. This demonstrates that our approach influences agent behavior and successfully discourages short cycles.  

\section{Conclusions}

The results of this study confirm that DRL is well-suited for economic dispatch problems involving uncertainty. Especially the more modern algorithms demonstrated higher sample efficiency and higher rewards. This applies to both continuous and discrete action selection. However, in different settings, for example if purchasing power from the grid is not possible and accurate load-following is required, continuous algorithms might be preferable.

The results further highlight the importance of an accurate O\&M cost calculation due to its impact on learned policies. This is particularly relevant where GTs are operated to balance the intermittency of RE. Here, the proposed dynamic O\&M scheme allows to better approximate the true cost of frequent but short cycles.

\begin{table}[hb]
	\begin{center}
		\caption{Accumulated rewards and sample efficiencies of tested algorithms.}\label{tb:performance}
		\begin{tabular}{c|c|c}
			Algorithm & Acc. reward & Sample efficiency \\\hline
			REINFORCE (continuous) & -4.22 & -5.27 \\
			REINFORCE (discrete) & -4.21 & -5.57 \\ 
			DQN & \textbf{-3.98} & -5.16 \\ 
			PPO & -4.01 & \textbf{-4.73} \\
			CEM & -4.40 & -6.37 \\ 
		\end{tabular}
	\end{center}
\end{table}

\begin{table}[h]
	\begin{center}
		\caption{Comparison of performance and policy characteristics for DQN and PPO with different O\&M scheme implementations.}
		\label{tb:results_overhaul}
		\begin{tabular}{ll|c|c|c}
			& &
			\begin{tabular}[c]{@{}c@{}}dynamic\\ O\&M\end{tabular} &
			\begin{tabular}[c]{@{}c@{}}hourly\\ O\&M\end{tabular} &
			\begin{tabular}[c]{@{}c@{}}no variable\\ O\&M\end{tabular} \\ \hline
			\multicolumn{1}{l|}{\multirow{2}{*}{\begin{tabular}[c]{@{}l@{}}Acc. reward\\ (M CAD)\end{tabular}}} 
			& DQN & -4.00 & -3.96 & -3.88 \\ \cline{2-5}
			\multicolumn{1}{l|}{}
			& PPO & -4.01 & -3.97 & -3.88 \\ \hline
			\multicolumn{1}{l|}{\multirow{2}{*}{\begin{tabular}[c]{@{}l@{}}GT oper.\\ hours\end{tabular}}}
			& DQN & 632 & 857 & 1129 \\ \cline{2-5} 
			\multicolumn{1}{l|}{}
			& PPO & 486 & 617 & 814 \\ \hline
			\multicolumn{1}{l|}{\multirow{2}{*}{\begin{tabular}[c]{@{}l@{}}GT oper.\\ cycles\end{tabular}}}
			& DQN & 117 & 247 & 394 \\ \cline{2-5} 
			\multicolumn{1}{l|}{}
			& PPO & 92 & 160 & 228                                                              
		\end{tabular}
	\end{center}
\end{table}

\begin{ack}
This work is supported by the McGill Engineering Doctoral Award (MEDA) and MITACS grant number IT13369. We thank Mahtab Salehii and Siemens Energy in Montreal for supporting our research.
\end{ack}

\bibliography{references}             % bib file to produce the bibliography
                                                     % with bibtex (preferred)
                                                   
%\appendix
%\section{A summary of Latin grammar}    % Each appendix must have a short title.
%\section{Some Latin vocabulary}              % Sections and subsections are supported  
%                                                                         % in the appendices.
\end{document}